\documentclass{article}

\usepackage[main, final, nonatbib]{neurips_2026}

\usepackage[utf8]{inputenc} 
\usepackage[T1]{fontenc}    
\usepackage{hyperref}       
\usepackage{url}            
\usepackage{booktabs}       
\usepackage{amsfonts}       
\usepackage{nicefrac}       
\usepackage{microtype}      
\usepackage{xcolor}         
\usepackage{amsmath} 
\usepackage{enumitem}
\usepackage{multirow}
\usepackage{graphicx}
\usepackage{algorithm}
\usepackage{algorithmic}
\usepackage{subcaption}
\usepackage{caption}

\usepackage{amsthm}  

\theoremstyle{definition}
\newtheorem{theorem}{Theorem}

\hypersetup{hidelinks} 

\title{CRESiST: Self-Enhancing Explorative Policy Learning for Simultaneous Speech Translation}

\author{\normalfont
  \textbf{Chenyang Le}$^{1}$\thanks{Work done during an internship at Alibaba Group.} \quad \textbf{Pei Zhang}$^{2,3}$ \quad \textbf{Yuxiang Chen}$^{4}$ \quad \textbf{Bing Han}$^{1,2}$ \\[2pt]
  \textbf{Derek F. Wong}$^{3}$ \quad \textbf{Baosong Yang}$^{2}$ \quad \textbf{Yanmin Qian}$^{1}$\thanks{Corresponding author.} \\[6pt]
  $^1$Shanghai Jiao Tong University \quad $^2$Alibaba Token Hub, Alibaba Group \\[1pt]
  $^3$University of Macau \quad $^4$University College London \\[4pt]
  \texttt{nethermanpro@sjtu.edu.cn} \quad \texttt{yanmin.qian@sjtu.edu.cn}
}

\begin{document}

{\renewcommand{\thefootnote}{\fnsymbol{footnote}}\maketitle}
\setcounter{footnote}{0}

\begin{abstract}
  Simultaneous Speech Translation (SimulST) requires balancing translation quality and latency. While Large Language Models (LLMs) excel in offline translation, current LLM-based SimulST systems predominantly rely on prefix-incremental generation. This paradigm suffers from redundant Key-Value cache recomputation and necessitates external Voice Activity Detection to process unbounded streams. To overcome these bottlenecks, we propose an end-to-end interleaved read-write framework built upon an omni-modal LLM. By interleaving speech chunks and text tokens, our architecture achieves full KV cache reuse, natively supporting segmentation-free, infinite-length inference. Furthermore, to learn optimal streaming policies without rigid offline alignments, we introduce a novel Actor-Critic self-enhancement pipeline. An RNN-T Actor dynamically samples read-write trajectories, while an LLM-based Critic evaluates these paths via intrinsic semantic and latency rewards, distilling the optimal policy back to the Actor. Extensive experiments on sentence-level (FLEURS, CoVoST2) and document-level (MuST-C, ACL 60/60) benchmarks demonstrate that our model establishes new state-of-the-art quality-latency Pareto frontiers, substantially outperforming existing cascaded and static-alignment baselines.
\end{abstract}

\section{Introduction}

Simultaneous Speech Translation (SimulST) aims to incrementally translate a continuous audio stream while minimizing delivery latency. Although Large Language Models (LLMs) have revolutionized offline translation, deploying them for unbounded real-time speech reveals a fundamental architectural conflict. State-of-the-art SimulST approaches \cite{ma2023Efficient, communication2023Seamless, lesimulmega, chen2024DivergenceGuided, hirschkind2026reina} predominantly rely on prefix-incremental generation \cite{liu2020low}, where newly arrived speech chunks are continuously appended to the historical prefix. This invalidates the Key-Value (KV) cache across incremental steps, causing a quadratic computational explosion and forcing systems to rely on external Segmentation, such as Voice Activity Detection (VAD). Such forced segmentation disrupts cross-sentence context, impairs coreference resolution, and fails to natively support infinite-length streams.

To bypass the KV cache bottleneck, recent studies explored interleaved read-write architectures \cite{ouyang2025infinisst, cheng2024towards} that append acoustic and textual tokens into a unified sequence. However, these models heavily depend on static, heuristic-driven word alignments (e.g., MFA\cite{mcauliffe2017montreal}) to construct supervised trajectories, failing to capture the dynamic, predictive nature of human interpreters. While some works adopt reinforcement learning \cite{xu2025seqpo, seed2025seed, labiausse2026hibikizero, ouyang2026hierarchical}, they still require offline aligned data and struggle to scale multilingually. To address this, we propose \textbf{CRESiST} (\textbf{C}ritic-\textbf{R}efined \textbf{E}xploratory \textbf{Si}multaneous \textbf{S}peech \textbf{T}ranslation), an interleaved framework that discovers its streaming policy without any external alignment supervision. Built on the Qwen2.5-omni-7B backbone \cite{xu2025qwen25omnitechnicalreport}, CRESiST discards prefix-incremental generation and segmentation, achieving full KV cache reuse. Crucially, it operates at a granular 400ms micro-chunk size, enabling a highly fine-grained streaming policy. 

To optimize this fine-grained policy without external alignments, we formulate the streaming decision process as a Markov Decision Process solved via a self-enhancing Actor-Critic pipeline. An auxiliary RNN-T acts as the Actor, actively exploring read-write trajectories. Simultaneously, the Speech LLM, equipped with an MoE Router and Refiner, serves as the Critic. The Critic evaluates trajectories using dynamic consistency and latency rewards, transferring its reward-maximizing preferences back to the Actor via reverse distillation. This iterative self-play allows CRESiST to autonomously discover adaptive, language-specific translation strategies (e.g., anticipating German verbs). Consequently, a single unified CRESiST model seamlessly translates Chinese and English streams into six diverse target languages.

Our main contributions are summarized as follows:
\begin{itemize}[leftmargin=0.5cm]
    \item \textbf{Unsupervised Streaming Policy in an Interleaved Framework:} While trained purely on offline parallel data, our framework autonomously transitions to an online interleaved paradigm, enabling full KV cache reuse for segmentation-free, infinite-length translation.
    \item \textbf{Fine-Grained Actor-Critic Policy Discovery:} We introduce a novel self-enhancement pipeline operating at a 400ms acoustic chunk size. Through reward-driven reverse distillation, the model autonomously learns optimal streaming policies without static forced alignments.
    \item \textbf{Multilingual Generalization:} A single unified CRESiST model supports Chinese/English inputs to six target languages, overcoming the limitations of language-pair-specific models.
    \item \textbf{State-of-the-Art Performance:} Extensive experiments across sentence-level and unbounded document-level benchmarks demonstrate that CRESiST outperforms strong baselines (e.g., InfiniSST), establishing new Quality-Latency Pareto frontiers.
\end{itemize}

\section{Related Work}

\paragraph{Read-Write Policy Learning}
Determining the optimal timing for translation is a core challenge in SimulST. A prevailing paradigm relies on static alignments from external tools (e.g., MFA) or probability mutations to construct supervised read-write trajectories \cite{ouyang2025infinisst, zhang2026direct, yang2026sasst, cheng2024towards, fu2026efficient, labiausse2025hibiki}. To circumvent the rigidity of forced alignments, fixed and heuristic approaches utilize continuous integrate-and-fire (CIF)\cite{dong2020cif}, wait-$k$ cascades, or attention boundary indicators \cite{deng2025simuls2s, guo2025streamuni, papi2023Attention, papi2023AlignAtt}. Recent unsupervised methods further explore dynamic boundary detection by leveraging offline-to-online inference divergence \cite{chen2024DivergenceGuided, hirschkind2026reina}, dual CIF distillation \cite{tan2026simulsense}, or implicit learning via mixture-of-experts \cite{lesimulmega}. Furthermore, to optimize the latency-quality trade-off beyond static supervision, advanced reinforcement learning techniques (such as PPO and GRPO) have been increasingly adopted to autonomously refine streaming translation strategies \cite{seed2025seed, ouyang2026hierarchical, labiausse2026hibikizero}.

\paragraph{Unbounded Audio Processing}
Processing continuous, infinite-length audio streams typically suffers from severe memory bottlenecks. Traditional systems heavily depend on external segmenters like VAD or acoustic heuristics \cite{tsiamas2022shas}, which inevitably disrupt cross-sentence contextual coherence. To mitigate this, recent studies have proposed history selection mechanisms that only retain fixed amounts of text tokens and their attention-covered acoustic frames \cite{papi2024streamatt}. Alternatively, formulating the continuous stream as multi-turn dialogues allows for the safe discarding of distant KV cache \cite{ouyang2025infinisst}; however, such formulations typically mandate tightly coupled supervised policies to function effectively.

\section{CRESiST}
\label{headings}

\subsection{Problem Formalization}
Unlike previous RL approaches in SimulST that couple translation timing with content generation—creating an intractable policy space—we decouple timing by formalizing the read-write strategy as a Markov Decision Process (MDP). Given an offline speech-text pair $\{\mathbf{S}, \mathbf{T}\}$ with acoustic chunks $\mathbf{S} = (s_1, \dots, s_{|\mathbf{S}|})$ and ground-truth text tokens $\mathbf{T} = (t_1, \dots, t_{|\mathbf{T}|})$, the MDP is defined by the quadruple $\langle \mathcal{S}, \mathcal{A}, \mathcal{P}, \mathcal{R} \rangle$. Specifically, the \textbf{state space ($\mathcal{S}$)} at step $k$ is $s_k = (\mathbf{S}_{\le u_k}, \mathbf{T}_{< v_k})$, representing the $u_k$ speech chunks read and $v_k$ target tokens written thus far. The \textbf{action space ($\mathcal{A}$)} is a binary space $\{\text{Read}, \text{Write}\}$ controlling the translation rhythm. The \textbf{transition dynamics ($\mathcal{P}$)} dictates the environment evolution $P(s_{k+1} \mid s_k, a_k)$: a \textit{Read} action appends the next speech chunk yielding $s_{k+1} = (\mathbf{S}_{\le u_k + 1}, \mathbf{T}_{< v_k})$, whereas a \textit{Write} action appends a text token to yield $s_{k+1} = (\mathbf{S}_{\le u_k}, \mathbf{T}_{< v_k + 1})$. Optimal write actions maximize congruence between \textbf{Data Dynamics} (ground-truth distribution $P_{data}(t_{v_k} \mid s_k)$) and \textbf{Model Dynamics} (LLM prediction $P_{model}(\cdot \mid s_k; \theta)$). Finally, the \textbf{reward function ($\mathcal{R}$)} provides a scalar $r_k(s_k, a_k)$ to maximize expected cumulative rewards, delicately balancing a \textit{dynamic consistency evaluation} (aligning the Model and Data dynamics) against a \textit{latency penalty} to ensure timely responses.

\begin{figure}
    \centering
    \includegraphics[width=1\linewidth]{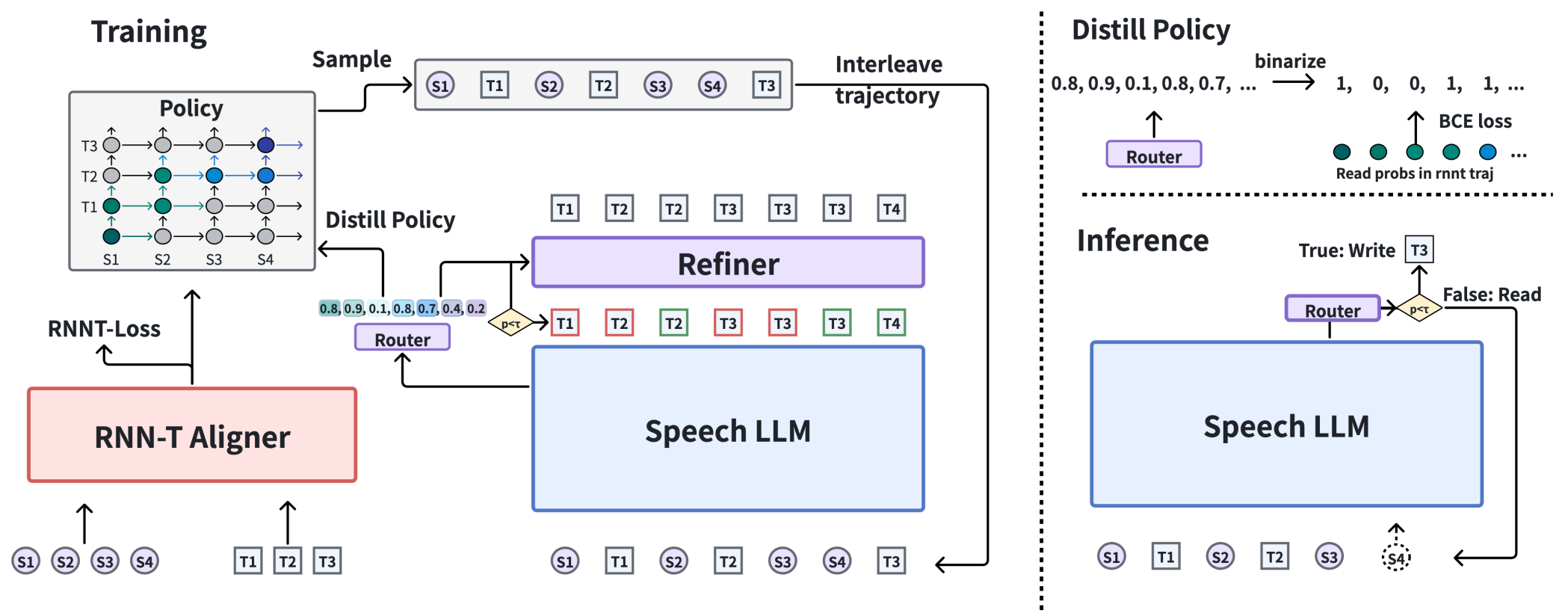}
    \caption{Training and inference diagram of CRESiST. In the interleaved sequence, red and green target tokens mark whether the Router score at that position passes the decision threshold, i.e., whether the LLM body loss $\mathcal{L}_{body}$ is computed there (tokens judged to be emitted too early are masked out). The Refiner loss $\mathcal{L}_{refiner}$ is computed at every text position, and the two losses are summed; the same ground-truth token is thus the target of two prediction heads over one sequence, not duplicated in it.}
    \label{fig:diagram}
\end{figure}

\subsection{Actor-Critic Training Framework}
\label{subsec:actor_critic}

To solve the MDP formulated in the previous section and transition toward an autonomous read-write interleaved paradigm, we propose a self-enhancing Actor-Critic training framework, outlined in Algorithm \ref{alg:train}. Given an offline speech-text corpus $\mathcal{D} = \{(\mathbf{S}, \mathbf{T})\}$, our framework jointly optimizes translation fidelity and the simultaneous streaming policy. By decoupling the generation process into an exploration-evaluation loop, we effectively compute the MDP transition dynamics and the reward function $\mathcal{R}$ without relying on heuristic alignments. 

\paragraph{The Actor for Trajectory Exploration.} We introduce an auxiliary RNN-T\cite{graves2012sequence} model parameterized by $\theta_{act}$ to serve as the Actor. Operating within the state space $\mathcal{S}$, the Actor explores the action space $\mathcal{A}$ by sampling a sequence of binary Read/Write decisions. For a given sample $(\mathbf{S}, \mathbf{T}) \sim \mathcal{D}$, it yields a trajectory of actions $\mathbf{a} \sim \pi_{act}(\cdot \mid \mathbf{S}, \mathbf{T}; \theta_{act})$, where $a_k \in \mathcal{V} \cup \{\phi\}$ (with $\phi$ denoting a ``Read'' action). This sampled trajectory simulates the transition dynamics $\mathcal{P}$, deterministically merging the source speech and target text into a unified, interleaved sequence $\mathbf{z} = \text{Interleave}(\mathbf{S}, \mathbf{T}, \mathbf{a})$. Through this exploration, the Actor constructs the streaming-style state transitions required by the downstream environment.

\paragraph{The Critic for Reward Evaluation and Validation.} The Qwen-omni LLM backbone, equipped with a Router-Refiner module and parameterized by $\theta_{cri}$, acts as the Critic. It ingests the state sequence $\mathbf{z}$ to evaluate the Actor's trajectory by computing the Reward Function ($\mathcal{R}$). The evaluation is twofold:
\begin{enumerate}[label=\arabic*), leftmargin=0.6cm]
    \item \textbf{Dynamic Consistency Reward ($\mathcal{R}_{consistency}$):} The Refiner and LLM body assess the congruence between Data Dynamics and Model Dynamics. By conditioning on local hidden states $\mathbf{H} = \text{LLM}(\mathbf{z}; \theta_{cri})$ and a globally compressed acoustic representation $\mathbf{h}_{global}$, it computes the token generation probabilities. We formally define the consistency reward as the negative log-likelihood of the generation targets:
    \begin{equation}
        \mathcal{R}_{consistency} = - (\mathcal{L}_{refiner} + \mathcal{L}_{body})
    \end{equation}
    where $\mathcal{L}_{refiner}$ and $\mathcal{L}_{body}$ are the cross-entropy generation losses of the Refiner and LLM body, respectively. High predictive accuracy yields a high consistency reward.
    \item \textbf{Latency Reward ($\mathcal{R}_{latency}$):} To ensure timely responses without relying on heuristic constraints, the Router module assesses the sequence of hidden states to predict the action probability $p_k = \text{Router}(\mathbf{H}_{k-1}; \theta_{cri})$ at each step, where $p_k \to 1$ signifies a \textit{Read} and $p_k \to 0$ signifies a \textit{Write}. We mathematically formalize the latency reward as the expected length of consecutive text generation in a given turn. Since the probability of executing at least $m$ consecutive writes starting from step $k$ is given by the cumulative product of write probabilities, the expected write length $\mathbb{E}[W_k]$ can be elegantly formulated as:
    \begin{equation}
        \mathcal{R}_{latency}^{(k)} = \mathbb{E}[W_k] = \sum_{m=1}^{M} \prod_{j=1}^{m} \big(1 - p_{k+j}\big)
    \end{equation}
    where $M$ denotes the maximum allowable length for a single conversation turn. By maximizing this expectation-based reward, the Critic intrinsically incentivizes the policy to emit longer textual chunks as early as possible, thereby minimizing the overall translation delay.
\end{enumerate}
Essentially, the Critic judges whether the state-action path traversed by the Actor maximizes the overall reward $\mathcal{R} = \mathcal{R}_{consistency} + \gamma \mathcal{R}_{latency}$, facilitating fluent generation with minimal delay.

\paragraph{Self-Enhancement via Reward-Driven Distillation.} The entire framework is optimized through an alternating closed-loop mechanism. First, the Critic is updated by minimizing the composite cost $\mathcal{J}_{critic}$, which integrates the negative rewards alongside an auxiliary load balancing loss ($\mathcal{L}_{balance}$) to stabilize the routing distribution:
\begin{equation}
    \mathcal{J}_{critic} = - \mathcal{R}_{consistency} - \alpha \mathcal{R}_{latency} + \beta \mathcal{L}_{balance}
\end{equation}
where $\alpha$ and $\beta$ scale the latency reward objective and the balancing penalty, respectively.

Subsequently, the Actor is updated by minimizing its own composite objective $\mathcal{J}_{actor}$. To emphasize the reward-based policy feedback, we position a novel \textit{Reverse Distillation} loss ($\mathcal{L}_{distill}$) as the primary optimization target, treating the standard RNN-T loss ($\mathcal{L}_{rnnt}$) as an auxiliary constraint to maintain basic acoustic-linguistic alignment:
\begin{equation}
    \mathcal{J}_{actor} = \mathcal{L}_{distill} + \lambda \mathcal{L}_{rnnt}
\end{equation}
where $\lambda$ acts as a down-scaling coefficient for the auxiliary loss. To compute the critical $\mathcal{L}_{distill}$, we use the reward-optimized Critic to binarize the Router's validated policy decisions via a predefined threshold $\tau$, constructing an optimal target action sequence $\hat{a}_k = \mathbb{I}(p_k \ge \tau)$. A Binary Cross-Entropy (BCE) distillation loss then forces the Actor to align its action probabilities with these LLM-preferred, reward-maximizing decisions:
\begin{equation}
    \mathcal{L}_{distill} = \sum_{k} \text{BCE} \Big(\hat{a}_k, \pi_{act}(a_k = \phi \mid \cdot; \theta_{act}) \Big)
\end{equation}

By continuously backpropagating gradients $\nabla_{\theta_{act}} \mathcal{J}_{actor}$, the RNN-T learns to autonomously sample trajectories that are intrinsically aligned with the LLM's optimal dynamics. We note that the pseudo-label $\hat{a}_k$ is treated with a stop-gradient on $\theta_{cri}$, decoupling the Actor update from gradient flow through the Critic. We run K = 8 Critic  steps followed by one Actor step for better convergence. Theoretical guarantees regarding the convergence of this joint optimization are provided in Appendix ~\ref{app:convergence}.

\subsection{Model Implementation}
\label{subsec:architecture}

\begin{figure}[htbp]
     \centering
     \begin{subfigure}[b]{0.45\textwidth}
         \centering
         \includegraphics[width=\textwidth]{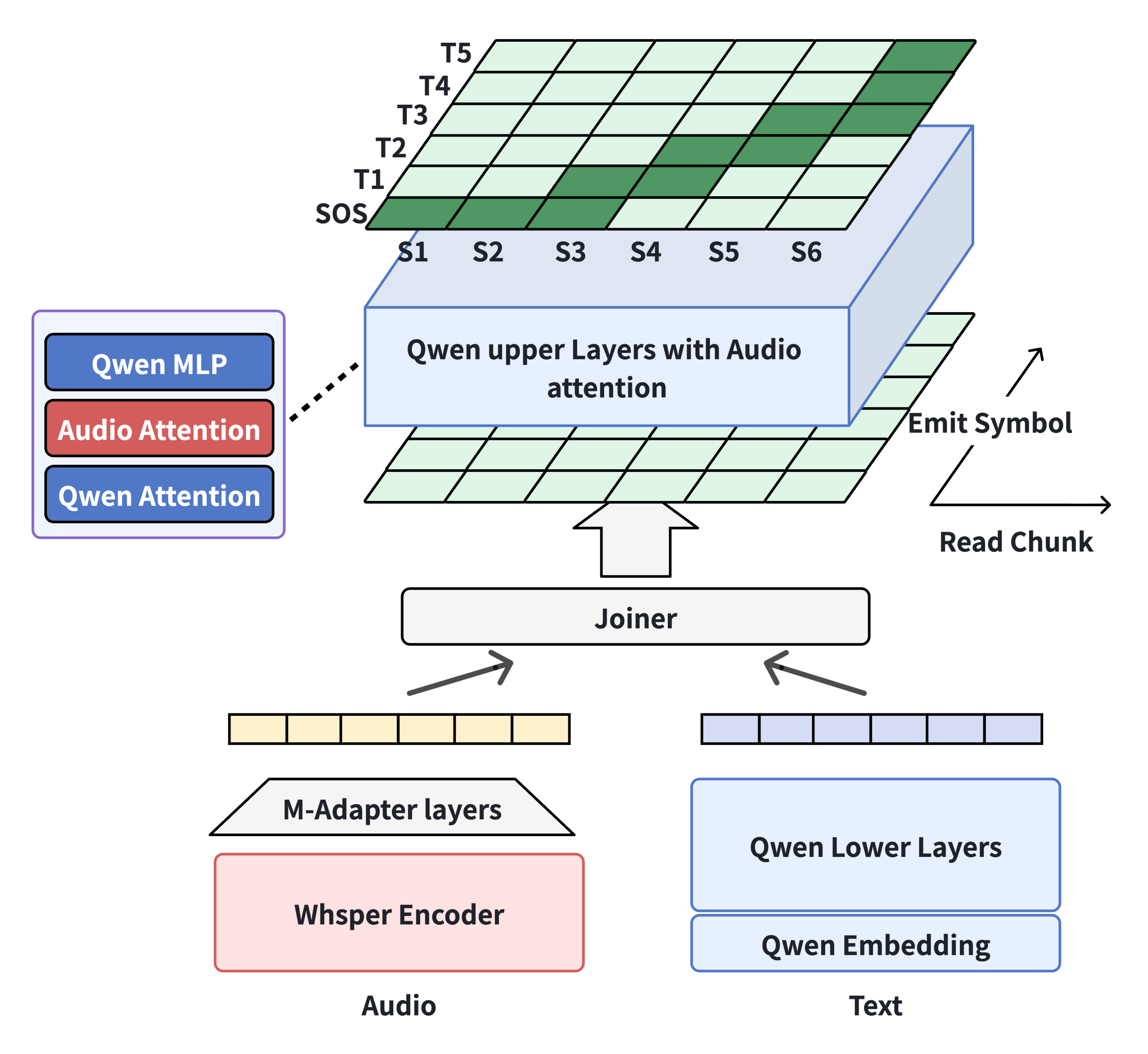}
         \caption{RNN-T Aligner}
         \label{fig:rnnt}
     \end{subfigure}
     \hfill 
     \begin{subfigure}[b]{0.42\textwidth}
         \centering
         \includegraphics[width=\textwidth]{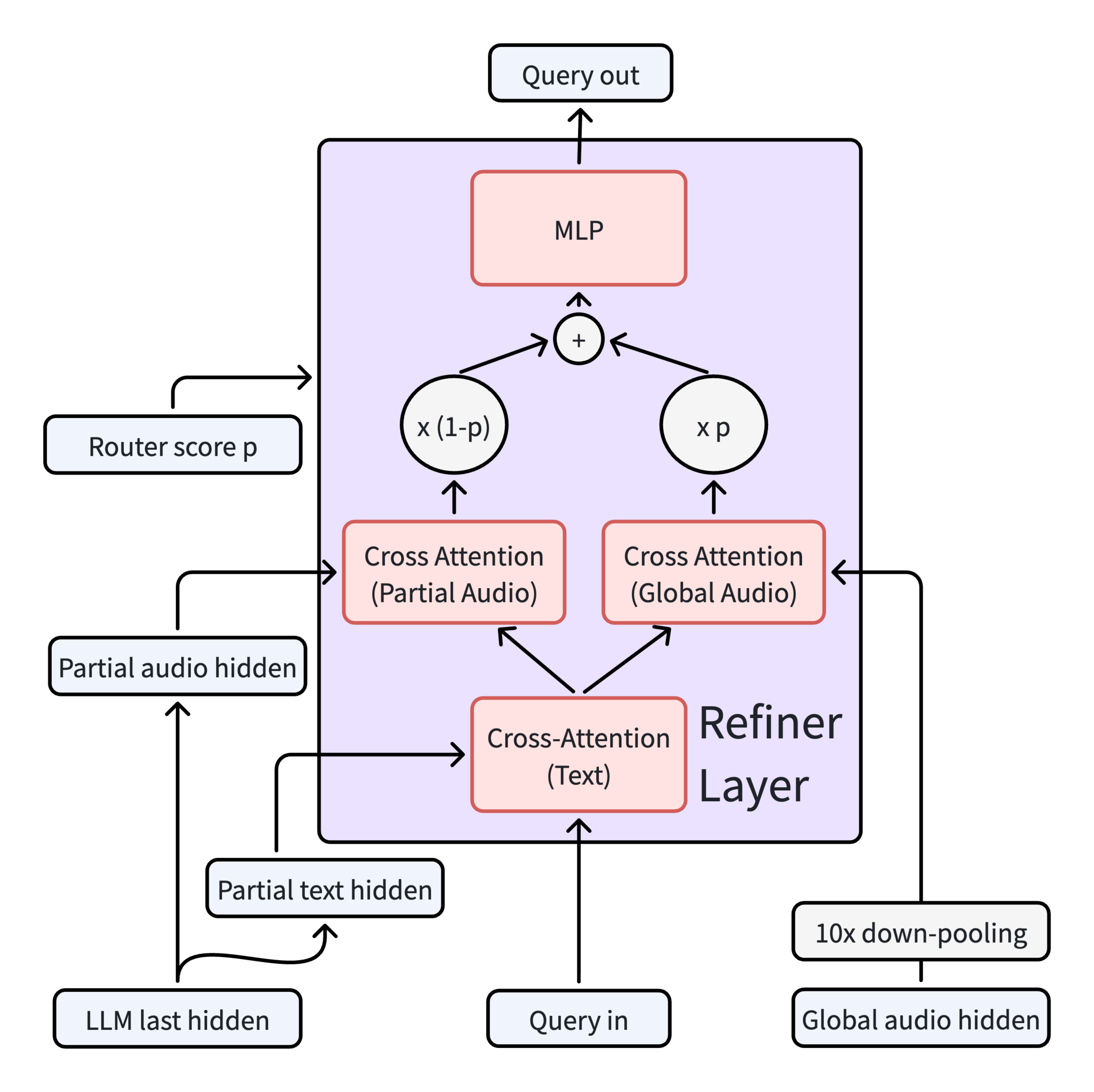}
         \caption{Refiner Layer}
         \label{fig:refiner}
     \end{subfigure}
     
     \caption{Structural details of the RNN-T aligner and the Refiner layer.}
     \label{fig:overall}
\end{figure}

To instantiate the Actor-Critic framework, the CRESiST architecture is bifurcated into two structural components: an RNN-T-based Actor for alignment exploration and an LLM-based Critic for validation and generation.

\subsubsection{The Actor Model: Alignment Explorer}
\label{subsubsec:actor_arch}

To avoid the computational inefficiency and "cold start" issues of rolling out interleaved sequences directly via the Speech LLM, the Actor is parameterized using an auxiliary Recurrent Neural Network Transducer (RNN-T). 

As illustrated in Figure \ref{fig:rnnt}, our Actor comprises an acoustic encoder, a text predictor, and a joint network, augmented with a 2D axial attention module. Each layer of this axial attention module consists of two causal self-attention mechanisms—operating sequentially along the text and speech axes—followed by an MLP, strengthening dual-axis modality interaction. 

For the specific configuration, the acoustic encoder is initialized with the pre-trained Whisper-medium encoder\cite{radford2023robust}, followed by two down-sampling M-Adapter layers to yield 2.5 Hz speech features (a 400ms window). A 400ms chunk-autoregressive mask is applied to strictly preserve causality. The text predictor and the axial attention module are initialized using the Qwen3-0.6B model, while the newly introduced audio-axis attention weights are initialized from scratch.

\paragraph{Read-Write Trajectory Construction.} 
To construct the interleaved sequence $\mathbf{z}$ defined in Section \ref{subsec:actor_critic}, we traverse the $(T \times U)$ alignment lattice. At each node $(t, u)$, the joint network outputs a blank (``Read'') emission probability $p_y^{(t,u)}$. The routing action $a_k$ is then directly sampled via:
\begin{equation}
    a_k \sim \text{Bernoulli}\big(p_y^{(t_k, u_k)}\big)
\end{equation}
The sampled path $\mathbf{a} = (a_1, \dots, a_K)$ governs the lattice coordinate updates and dictates the exact merge boundaries for the LLM input space. Note that every sampled trajectory is a monotonic path on the lattice from $(0, 0)$ to $(T, U)$: a blank advances only along the audio axis and a token emission only along the text axis, and once the last audio chunk is reached all remaining target tokens are emitted. Every trajectory therefore contains exactly $U$ Write actions, so no target token is ever dropped from $\mathbf{z}$.

\subsubsection{The Critic Model: Speech LLM and Validator}
\label{subsubsec:critic_arch}

The Critic evaluates the Actor's trajectories and generates the final translation. It is built upon a Qwen-omni Large Language Model backbone, structurally augmented with Router and Refiner modules.

\paragraph{Speech LLM Backbone.} 
Adapted from the ``thinker'' component of the Qwen2.5-omni-7B, the speech encoder is modified for streaming by applying a windowed chunk autoregressive mask (400ms chunk size, 20-chunk local attention window, yielding an 8-second receptive field). Cyclic sinusoidal positional encodings are applied every 2 seconds. 

The input is formatted using a multi-turn dialogue template. Task-level directives are embedded in the system prompt, and each audio chunk with its interleaved textual actions serves as a single conversational turn. The backbone processes this to yield local hidden states $\mathbf{H} = \text{LLM}(\mathbf{z}; \theta_{cri})$ and a globally compressed acoustic representation $\mathbf{h}_{global} = \text{AvePool}(\text{LLM}(\mathbf{S}; \theta_{cri}))$.

\paragraph{Router Module.}
Implemented as a lightweight 2-layer MLP with a Sigmoid activation, the Router projects the preceding hidden state $\mathbf{H}_{k-1}$ to a scalar probability $p_k \in [0, 1]$. While Section \ref{subsec:actor_critic} details its role in the latency reward ($\mathcal{R}_{latency}$), structurally, $p_k$ is thresholded by a predefined $\tau$ during inference to trigger either a ``Read'' ($p_k \ge \tau$) or ``Write'' ($p_k < \tau$) action.

\paragraph{Refiner Module.}
We adopt the Refiner architecture \cite{lesimulmega} for final generation. As shown in Figure \ref{fig:refiner}, the core structural integration involves repurposing the Router's probability output $p$ as a dynamic routing scalar. The Refiner directly scales the global acoustic gain by $p$ and the local acoustic gain by $(1-p)$. This self-adaptive gating synchronizes the generation context with the streaming timing state before the refined features are projected to compute the consistency reward ($\mathcal{R}_{consistency}$). During inference, global acoustic information is absent. The Router can make policies without global information, and we do not use the Refiner at inference.

\section{Experiment and Results}
\label{others}

\subsection{Experimental Setup}
\label{sec:exp_setup}

\paragraph{Training Data}
\label{subsec:training_data}
To support multilingual translation, we aggregate three open-source ASR datasets: Emilia\cite{emilia}, WenetSpeech\cite{zhang2022wenetspeech}, and GigaSpeech\cite{chen2021gigaspeech}. We utilize the Qwen3-235B-A22B\footnote{\url{https://huggingface.co/Qwen/Qwen3-235B-A22B-Instruct-2507}} model to generate pseudo-labels for the transcribed texts, translating them into Chinese, English, and the other 4 European languages (DE, FR, ES, IT). (Chinese and English account for more than 50\% of the pseudo-labels). We augment WenetSpeech and GigaSpeech by concatenating consecutive clips into long segments of a maximum of 30 seconds. All training stages dynamically sample from this unified corpus for training and validation.

\paragraph{Training Details}
\label{subsec:training_details}
Our training pipeline is implemented using PyTorch Lightning. The Speech LLM is parameter-efficiently fine-tuned using LoRA\cite{hu2021lora} with a rank of $r=16$. When training the LLM for SimulST task, we also train offline ST and MT tasks in a multi-task way to preserve offline performance.

Following our proposed Actor-Critic paradigm, the training strictly proceeds in three stages:
\begin{enumerate}[label=\arabic*), leftmargin=0.6cm]
    \item \textbf{Actor Pre-training:} The auxiliary RNN-T Aligner is trained with the standard RNN-T loss to establish basic acoustic-linguistic alignments. Only the alignment task itself is new at this stage: the acoustic encoder is initialized from Whisper-medium, and the text-axis attention and MLP of each axial layer inherit the corresponding Qwen3-0.6B layer (the bottom 12 layers are frozen and receive no audio-axis attention; the upper 16 layers are augmented), while the audio-axis attention and the joint network are randomly initialized.
    \item \textbf{Critic Pre-training:} Freezing the RNN-T parameters, we use its sampled trajectories to construct interleaved sequences. The Speech LLM (including the Router and Refiner) is optimized to adapt to this streaming input format.
    \item \textbf{Joint Self-Enhancement:} All modules are unfrozen. The framework undergoes joint training via reverse distillation (as detailed in Section \ref{subsec:actor_critic}), continuously refining the Actor's sampling policy alongside the Critic's generation capability.
\end{enumerate}

\paragraph{Benchmarks.} We evaluate our model across diverse streaming scenarios. For \textbf{sentence-level evaluation}, we utilize the FLEURS\cite{conneau2023fleurs} and CoVoST 2\cite{wang2021covost} datasets across the 10 supported directions. Notably, on FLEURS, we decouple the evaluation into Chinese-English bidirectional translations (to ensure a direct comparison with Seed-LiveInterpret2) and the remaining 8 directions. For \textbf{document-level continuous streaming}, we utilize the MuST-C and ACL6060 test sets (En-Zh, En-De). We use OmniSTEval for long-form evaluation.\footnote{\url{https://github.com/pe-trik/OmniSTEval/tree/main}}

\paragraph{Baselines.} We benchmark our framework against strong state-of-the-art systems, including open-source foundations (\textbf{Seamless Streaming}), commercial SOTA \textbf{Seed-LiveInterpret2}, and academic simultaneous models (\textbf{AlignATT}, \textbf{StreamUni}, \textbf{InfiniSST}). More details about the baseline implementation can be found in Appendix \ref{sec:adibaseline}.

\paragraph{Metrics.} Simultaneous translation requires balancing translation fidelity and streaming latency. For Translation Quality, we utilize \textbf{XCOMET-XXL}\footnote{\url{https://huggingface.co/Unbabel/XCOMET-XXL}}, a neural metric that aligns significantly better with human judgments compared to traditional n-gram metrics\cite{guerreiro-etal-2024-xcomet}. For \textbf{Streaming Latency}, we apply input-adaptive metrics to avoid heuristic bias: we report the Length-Adaptive Acoustic Latency (\textbf{LAAL})\cite{papi-etal-2022-generation} for sentence-level benchmarks, and adopt the recently proposed \textbf{LongYAAL} metric for document-level evaluations to accurately capture the cumulative delays in continuous long-form audio streams\cite{polak2025better}.

\subsection{Main Results}

\begin{figure}[tbp]
    \centering
    \includegraphics[width=1\linewidth]{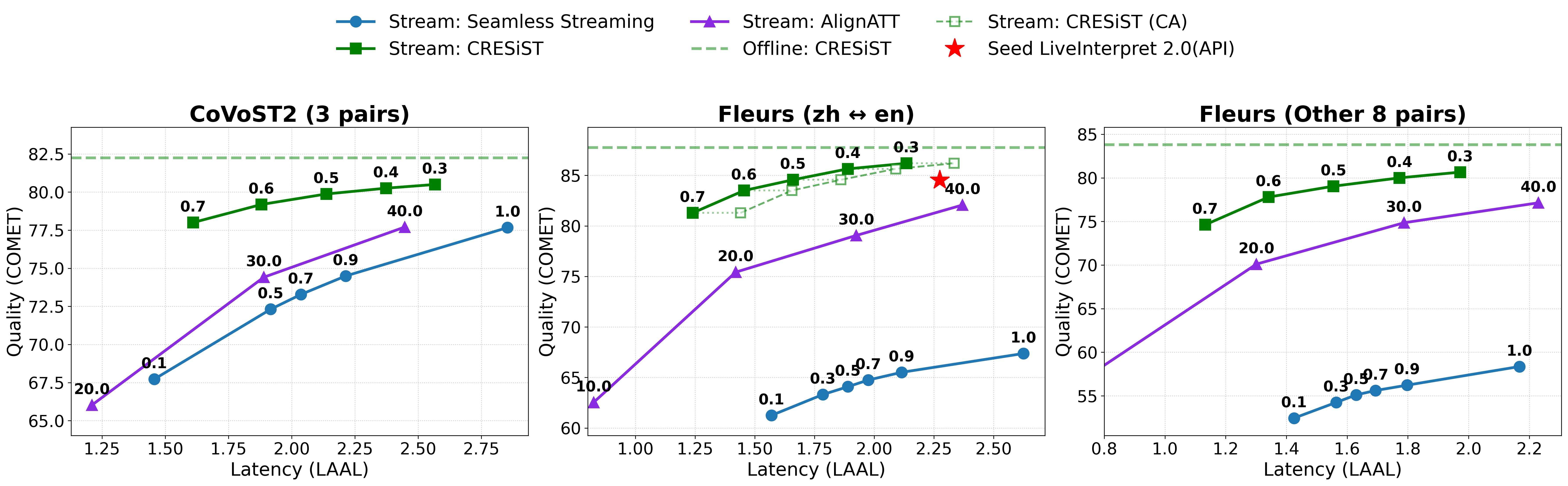}
    \caption{Sentence-level quality-latency trade-off curves on the CoVoST2 and FLEURS datasets. The proposed framework consistently establishes a superior Pareto frontier compared to Seamless Streaming and AlignATT across all sub-tasks. }
    \label{fig:placeholder}
\end{figure}

\paragraph{Sentence-Level Quality-Latency Trade-offs.}
Figure \ref{fig:placeholder} illustrates the quality-latency Pareto frontiers on the CoVoST2 and FLEURS benchmarks. By varying the decision threshold (e.g., $\tau \in [0.3, 0.7]$), we plot the translation quality against the latency metric. Our proposed framework consistently establishes a new state-of-the-art frontier, dominating both Seamless Streaming and AlignATT across all latency regimes. Notably, on the FLEURS (zh $\leftrightarrow$ en) benchmark, our streaming model strictly outperforms the commercial Seed LiveInterpret 2.0 API, achieving higher translation quality while maintaining a comparable or superior latency profile. Furthermore, the curve demonstrates that our model degrades gracefully at extremely low latency requirements (e.g., LAAL $< 1.5$), proving the effectiveness of the reverse-distilled streaming policy.

\paragraph{Document-Level Continuous Streaming.}
Evaluating on isolated sentences often obscures the compounding errors inherent in real-world, long-form translation. Table \ref{tab:doc_results} presents our results on the MuST-C and ACL 60/60 benchmarks across three settings: \textit{Offline w/ Oracle Segmentation}, \textit{Streaming w/ Oracle Segmentation}, and the most rigorous \textit{Streaming w/o Segmentation}. The contrast between the latter two settings effectively reveals a model's true capacity to process continuous, unbounded audio streams.

Notably, while InfiniSST is explicitly designed for infinite audio streams, it exhibits a paradoxical performance drop when evaluated at the sentence level with oracle boundaries. In contrast, CRESiST delivers consistently high performance across both streaming scenarios. Strikingly, our unsegmented streaming performance closely trails the oracle segmentation with a marginal degradation of only 1–2 COMET points, demonstrating that CRESiST can flawlessly manage infinite context without relying on explicit segmentation.

Furthermore, we observe that offline incremental methods such as StreamATT serve as highly competitive baselines on certain language pairs. However, their performance is notably inconsistent; on other language pairs, historical context selection leads to a catastrophic drop in translation quality (e.g., on the ACL 60/60 en-de dataset, the COMET score plummets from 80.83 under oracle segmentation to 70.61) coupled with severe latency inflation. Ultimately, CRESiST successfully navigates infinite audio streams, avoiding the pitfalls of historical amnesia and alignment drift while maintaining state-of-the-art translation quality and minimal continuous delay. CRESiST is also complementary to RL-based policy optimization: GRPO post-training further improves it, whereas the same recipe cannot close the gap from a direct-SFT base (Appendix~\ref{app:grpo}).

\begin{table}[htbp]
\centering
\setlength{\tabcolsep}{3pt}
\caption{Document-level simultaneous translation performance on the MuST-C and ACL 60/60 benchmarks. The COMET is evaluated through XCOMET-XXL. Our proposed CRESiST model achieves state-of-the-art translation quality with minimal continuous delay, demonstrating exceptional robustness in the challenging unsegmented streaming scenario. \textit{Oracle Segmentation} denotes sentence-level segmentation taken from the native sentence timestamps of the test sets, i.e., inference is run independently on each human-segmented sentence.}
\label{tab:doc_results}
\resizebox{\textwidth}{!}{
\begin{tabular}{l cc cc cc cc}
\toprule
 & \multicolumn{4}{c}{MuST-C} & \multicolumn{4}{c}{ACL 60/60} \\
\cmidrule(lr){2-5} \cmidrule(lr){6-9}
 & \multicolumn{2}{c}{en-zh} & \multicolumn{2}{c}{en-de} & \multicolumn{2}{c}{en-zh} & \multicolumn{2}{c}{en-de} \\
\cmidrule(lr){2-3} \cmidrule(lr){4-5} \cmidrule(lr){6-7} \cmidrule(lr){8-9}
Method & COMET & LongYAAL & COMET & LongYAAL & COMET & LongYAAL & COMET & LongYAAL \\
\midrule
\multicolumn{9}{c}{\textit{Offline w/ Oracle Segmentation}} \\ 
Qwen2.5-Omni & 79.41  & / & 91.47  & / &  84.42 & / & 91.72  & / \\
CRESiST & 79.11 & / & 91.35 & / & 83.36 & / & 90.03 & / \\
\midrule \addlinespace[0.5ex]
\multicolumn{9}{c}{\textit{Streaming w/ Oracle Segmentation}} \\
InfiniSST (m=3) & 68.46 & 2.18 & 79.22 & 1.93 & 59.71 & 2.17 & 65.09 & 2.17 \\
AlignATT ($f$=30) & 73.45 & 1.39 & 84.10 & 1.10 & 76.94 & 1.74 & 80.83 & 1.57 \\
CRESiST ($\tau$=0.5) & \textbf{75.30} & \textbf{1.29} & \textbf{87.75} & \textbf{0.80} & \textbf{80.06} & \textbf{1.49} & \textbf{87.23} & \textbf{1.00} \\
\midrule \addlinespace[0.5ex]
\multicolumn{9}{c}{\textit{Streaming w/o Segmentation}} \\
StreamUni (k=3) & 52.96 & 2.91 & 73.45 & 2.01 & 41.90 & 1.64 & 71.03 & 1.64 \\ 
InfiniSST (m=3) & 70.66 & 1.88 & 78.74 & 1.46 & 68.62 & 1.80 & 63.52 & 1.46 \\
StreamATT ($f$=30) & 70.23 & 2.05 & 74.35 & 2.42 & 76.63 & 2.15 & 70.61 & 1.89 \\
CRESiST ($\tau$=0.5) & \textbf{73.73} & \textbf{1.14} & \textbf{85.96} & \textbf{0.76} & \textbf{79.45} & \textbf{1.49} & \textbf{85.59} & \textbf{1.08} \\
\bottomrule
\end{tabular}
}
\end{table}

\section{Ablation Studies}
\label{sec:ablations}

\subsection{Effect of Training Strategies}
\begin{figure}[htbp]
    \centering
    \includegraphics[width=1\linewidth]{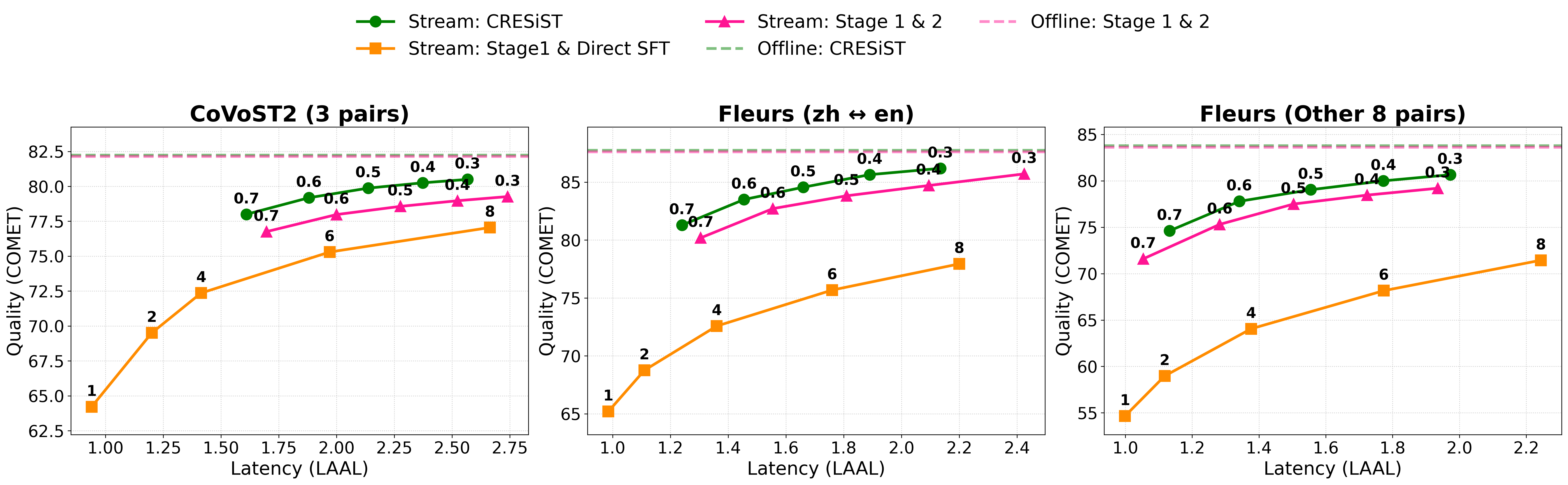}
    \caption{Ablation on training strategies. ``Direct SFT'' uses RNN-T aligned data directly (similar to InfiniSST). ``Stage 1 \& 2'' represents pipeline training without the final joint optimization. ``CRESiST'' indicates our full joint training framework, which yields the most significant latency reduction, particularly on resource-rich directions (e.g., zh $\leftrightarrow$ en).}
    \label{fig:ablation_train}
\end{figure}

We first investigate the impact of our self-enhancing Actor-Critic training paradigm. Figure \ref{fig:ablation_train} presents the quality-latency trade-offs under different training strategies. For fairness, the training steps are kept identical. The baseline (``Stage 1 \& Direct SFT''), which directly utilizes RNN-T aligned data for Supervised Fine-Tuning akin to the approach in InfiniSST, yields the poorest performance. It struggles to run at a granularity of 400 milliseconds (latency multiplier=1). The pipeline approach (``Stage 1 \& 2''), which trains the modules sequentially without the final joint optimization, shows noticeable improvements over the direct SFT. 

In contrast, our full CRESiST framework with joint self-enhancement achieves the best overall Pareto frontier. The improvement is particularly pronounced on the Chinese-English (zh $\leftrightarrow$ en) pairs, where the overall latency is significantly reduced while maintaining high translation quality. 

\vspace{-5mm}
\subsection{Sensitivity to the Latency Penalty Weight}
Next, as a sensitivity analysis rather than an ablation, we examine the impact of the latency penalty hyperparameter $\alpha$ on the system, with all variants trained for an identical number of steps. As illustrated in Figure \ref{fig:ablation_hyper}, $\alpha$ explicitly governs the trade-off between translation quality and streaming latency. As expected, when $\alpha$ is large, the model aggressively penalizes delays, resulting in a compressed, lower overall latency profile, but at the cost of relatively degraded translation quality. Conversely, an excessively small $\alpha$ fails to adequately constrain the streaming policy, leading to disproportionately high latency spikes at specific decision thresholds.

\subsection{Actor Architecture: Axial Blocks vs. Conventional RNN-T}
Finally, we evaluate the architectural design of our auxiliary RNN-T Actor by comparing our proposed model (incorporating 2D axial attention blocks) against a conventional RNN-T architecture. We assess the alignment quality on the validation set using two metrics: the perplexity (PPL, base $e$) of the emitted target text and the Average Lagging (AL, measured in seconds) of the generated interleaved sequence. \cite{maetal2019stacl}

\begin{figure}[htbp]
    \centering
    
    \begin{minipage}[c]{0.45\textwidth}
        \centering
        \includegraphics[width=\textwidth]{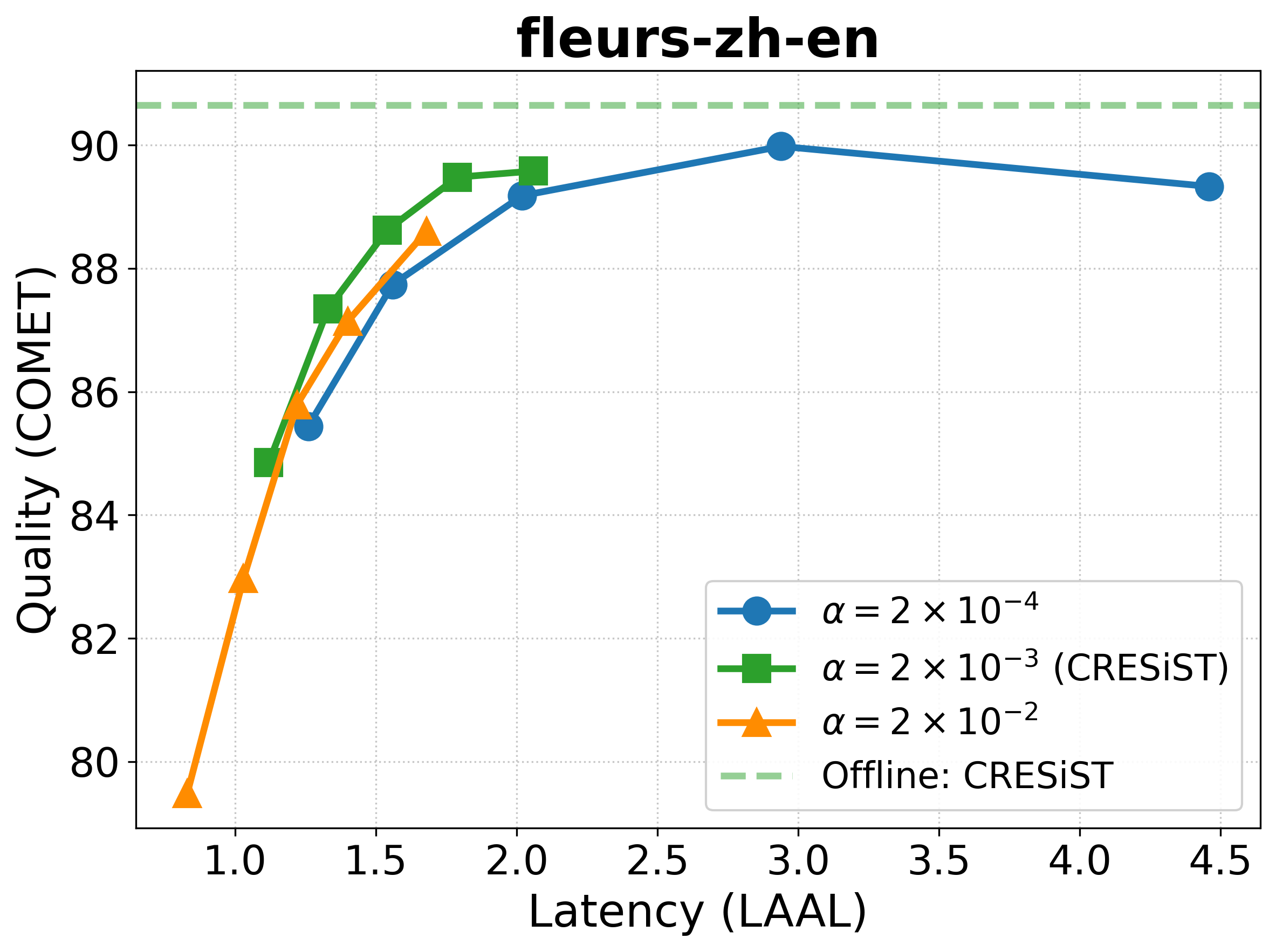}
        \captionof{figure}{Impact of the latency penalty weight on the quality-latency trade-off. Larger weights aggressively compress latency at the expense of quality, while excessively small weights lead to prolonged delays at specific thresholds.}
        \label{fig:ablation_hyper}
    \end{minipage}
    \hfill 
    \begin{minipage}[c]{0.45\textwidth}
        \centering
        \captionof{table}{Comparison of the Actor architecture using a conventional RNN-T versus our proposed model with axial blocks. Evaluated on the validation set by text perplexity (PPL) and Average Lagging (AL, in seconds). $N_{pd}$ and $N_{ax}$ are the number of layers in the predictor and the Axial attention module respectively}
        \label{tab:ablation_actor}
        \begin{tabular}{lcccc}
            \toprule
            Model & $N_{pd}$ & $N_{ax}$ & PPL & AL \\
            \midrule
            Conventional & 28 & 0 & 22.3 & 8.03\\
            Proposed & 12 & 16 & \textbf{6.03} & \textbf{1.81}\\
            \bottomrule
        \end{tabular}
    \end{minipage}
    
\end{figure}

As detailed in Table \ref{tab:ablation_actor}, both the two checkpoints are trained for 600k steps. The conventional RNN-T struggles significantly to produce optimal read-write paths, exhibiting abnormally high latency (AL = 8.03s) and poor text generation confidence (PPL = 22.3). The inclusion of axial blocks effectively strengthens the interaction between audio and text modalities, drastically improving alignment behavior. This design successfully drops the PPL to 6.03 and reduces the average lagging to 1.81s, providing a much higher-quality interleaved sequence for the downstream Speech LLM. We further verify in Appendix~\ref{app:rnnt_infer} that the Actor is effective as a policy explorer but not as the final translator: reverse distillation transfers its latency behavior, while its translation quality stays far below the Speech LLM.

\section{Conclusion and Limitations}
\label{sec:conclusion}

\paragraph{Conclusion.} 
In this work, we introduced a novel self-enhancing Actor-Critic framework designed to transition simultaneous speech translation from rigid, incremental processing to a dynamic, read-write interleaved paradigm. By decoupling policy exploration and reward evaluation, our method empowers an auxiliary RNN-T Actor to construct optimal streaming trajectories, while the Speech LLM Critic leverages its Router-Refiner module to ensure high-fidelity generation. Through a closed-loop reverse distillation mechanism, the framework autonomously aligns the streaming policy with the LLM's generation rhythm. Extensive experiments demonstrate that our approach establishes new state-of-the-art quality-latency trade-offs across both sentence-level and challenging document-level benchmarks.

\paragraph{Limitations and Future Work.} 
Despite the promising results, our framework exhibits a few limitations that warrant future investigation:

\begin{itemize}[leftmargin=0.4cm]
    \item \textbf{Data and Domain Bottlenecks:} As this work primarily focuses on streaming policy optimization, we didn't perform rigorous data engineering. The absolute translation quality is fundamentally bottlenecked by the backbone capacity and offline SFT data, which sub-optimally affect overall performance. For example, we observed an offline performance drop on the ACL60/60 dev set compared to the vanilla Qwen2.5-omni model, indicating domain distribution discrepancies and catastrophic forgetting in specialized academic domains.
    
    \item \textbf{Document-level Latency Control:} While adjusting the decision threshold ($\tau$) effectively modulates the sentence-level quality-latency trade-off, it yields marginal tuning effects during document-level continuous streaming. Regulating the compounding dynamics of infinite audio streams likely requires dynamic penalty mechanisms, such as the context-aware latency multiplier utilized in InfiniSST, rather than relying solely on a static probability threshold.
\end{itemize}

\begin{ack}
This work was supported in part by China NSFC project under Grants No. 62631016, No. 62601524 and No. U25A20409; in part by SJTU Med-X (Medicine \& Engineering) Translational Research Grant (YG2025LC09); in part by the UM and UMDF (Grant Nos. MYRG-GRG2024-00165-FST-UMDF, MYRG-GRG2025-00236-FST, MYRG-GRG2026-00018-FIC, MYRG-CRG2026-00007-FIC); and in part by the Alibaba Innovative Research Program.
\end{ack}

\medskip

\small
\bibliographystyle{IEEEtran}
\bibliography{mybib}

\newpage

\appendix

\section{Overall Training Algorithm}

Algorithm \ref{alg:train} summarizes the complete end-to-end training procedure of our proposed Actor-Critic framework for simultaneous translation.

\begin{algorithm}[htbp]
\caption{Actor-Critic Training for Simultaneous Translation}
\label{alg:train}
\textbf{Input:} Dataset $\mathcal{D} = \{(\mathbf{S}, \mathbf{T})\}$, Actor parameters $\theta_{act}$, Critic parameters $\theta_{cri}$, Learning rate $\eta$, Decision threshold $\tau$, Loss weights $\lambda, \alpha, \beta$, critic step num K. 
\begin{algorithmic}[1]
\WHILE{not converged}
    \STATE Sample a batch of speech-text pairs: $(\mathbf{S}, \mathbf{T}) \sim \mathcal{D}$
    
    \vspace{1.5mm}
    \STATE \textit{\textbf{// Step 1: Actor Exploration \& Transition Dynamics ($\mathcal{S}, \mathcal{A}, \mathcal{P}$)}}
    \STATE Sample alignment path ($a_k \in \mathcal{V} \cup \{\phi\}$, $\phi$ denotes Read): $\mathbf{a} \sim \pi_{act}(\cdot \mid \mathbf{S}, \mathbf{T}; \theta_{act})$
    \STATE Construct interleaved state sequence: $\mathbf{z} = \text{Interleave}(\mathbf{S}, \mathbf{T}, \mathbf{a})$
    
    \vspace{1.5mm}
    \STATE \textbf{\textit{// Step 2: Critic Forward Pass \& State Evaluation}}
    \STATE Extract main body hidden states: $\mathbf{H} = \text{LLM}(\mathbf{z}; \theta_{cri})$
    \STATE Predict router probabilities ($p_k \in [0, 1]$): $p_k = \text{Router}(\mathbf{H}_{k-1}; \theta_{cri})$
    \STATE Extract global acoustic state: $\mathbf{h}_{global} \leftarrow \text{AvePool}(\text{LLM}(\mathbf{S}; \theta_{cri}))$
    \STATE Predict refiner distributions: $P_{refiner}(t_i \mid \mathbf{z}_{\le \text{idx}(i)}, \mathbf{H}, \mathbf{h}_{global}; \theta_{cri})$
    
\vspace{1.5mm}
    \STATE \textbf{\textit{// Step 3: Compute Reward Function $\mathcal{R}$ \& Critic Loss}}
    \STATE Compute Consistency Reward: 
    \STATE $\mathcal{R}_{consistency} \leftarrow \sum_{i} \log P_{refiner}(t_i \mid \cdot) + \sum_{\{k \mid p_k < \tau\}} \log P_{llm}(y_k \mid \mathbf{H}_k)$
    \STATE Compute Latency Reward: $\mathcal{R}_{latency} \leftarrow \sum_{k} \sum_{m=1}^{M} \prod_{j=1}^{m} \big(1 - p_{k+j}\big)$
    \STATE Compute Load Balancing constraint: $\mathcal{L}_{balance}(p_k)$
    \STATE Aggregate Critic cost: $\mathcal{J}_{critic} = - \mathcal{R}_{consistency} - \alpha \mathcal{R}_{latency} + \beta \mathcal{L}_{balance}$
    
    \vspace{1.5mm}
    \STATE \textbf{\textit{// Step 4: Policy Feedback \& Actor Optimization}}
    \STATE Compute base Actor loss: $\mathcal{L}_{rnnt} = -\log \pi_{act}(\mathbf{T} \mid \mathbf{S}; \theta_{act})$
    \STATE Derive optimal action targets from Critic: $\hat{a}_k = \mathbb{I}(p_k \ge \tau)$
    \STATE Compute policy distillation loss: $\mathcal{L}_{distill} = \sum_{k} \text{BCE} \Big(\hat{a}_k, \pi_{act}(a_k = \phi \mid \cdot; \theta_{act}) \Big)$
    
    \vspace{1.5mm}
    \STATE \textit{\textbf{// Step 5: Parameter Updates}}
    \IF{$t \bmod (K+1) = 0$}
        \STATE Update Actor parameters: $\theta_{act} \leftarrow \theta_{act} - \eta \nabla_{\theta_{act}} \big( \mathcal{L}_{distill} + \lambda \mathcal{L}_{rnnt} \big)$
    \ELSE
        \STATE Update Critic parameters: $\theta_{cri} \leftarrow \theta_{cri} - \eta \nabla_{\theta_{cri}} \mathcal{J}_{critic}$
    \ENDIF
\ENDWHILE
\end{algorithmic}
\end{algorithm}

\section{Convergence Analysis of the Reverse Distillation Loop}
\label{app:convergence}

This appendix provides a local convergence analysis for the Actor–Critic reverse distillation loop in Section~\ref{subsec:actor_critic}, formalizing the intuition that the Actor's quality is controlled by the Critic's quality despite the self-referential pseudo-labels. The analysis is local; we do not claim global convergence on a non-convex landscape.

We re-parameterize the Actor and Critic as $\theta$ ($\theta_{act}$) and $\phi$ ($\theta_{cri}$) for analytical convenience. The two-timescale update at outer step $t$ runs $K=8$ Critic SGD steps followed by one Actor step; the pseudo-label $\hat{a}_k = \mathbb{I}(\mathrm{sg}[p_k(\phi)] \ge \tau)$ uses a stop-gradient on $\phi$. Throughout, we assume standard regularity in a local neighborhood $\mathcal{N}\times\mathcal{M}$ of an equilibrium $(\theta^*,\phi^*)$: local smoothness of $J_a$ and $J_c$ in their first arguments, a quadratic-growth condition for the Critic around $\phi^*(\theta) := \arg\min_\phi J_c(\phi;\theta)$, cross-Lipschitzness of $\nabla_\phi J_c$ in $\theta$, a Tsybakov-style margin condition $\mathbb{P}(|p(s;\phi)-\tau|\le\delta) \le \kappa\delta$ on the Router output, a Polyak–\L{}ojasiewicz condition for $J_a$, unbiased minibatch gradients with bounded variance $\sigma^2$, and bounded gradients on $\mathcal{N}\times\mathcal{M}$. These conditions are weaker than those typically invoked in the analysis of two-timescale stochastic approximation (Borkar 2008; Karimi et al.~2016) and are widely satisfied by overparameterized neural networks. We refer the reader to standard references for their formulation.

\begin{theorem}[Local linear convergence to a noise-controlled neighborhood]
\label{thm:main}
Let $e_t := \phi_{t+1}-\phi^*(\theta_t)$ denote the Critic tracking error and $V_t := J_a(\theta_t;\phi^*(\theta_t)) - \min_\theta J_a(\theta;\phi^*(\theta_t))$ the Actor optimality gap. Under the regularity conditions above, there exist constants $\eta_a^0, \eta_c^0, K_0, c, C_\sigma > 0$ and a contraction rate $\rho^*\in(0,1)$ such that for any
\begin{equation*}
\eta_c \le \eta_c^0, \qquad \eta_a \le \eta_a^0, \qquad K \ge K_0,
\end{equation*}
and any initialization $(\theta_0,\phi_0) \in \mathcal{N}\times\mathcal{M}$, the Lyapunov function $\Phi_t := V_t + c\,\mathbb{E}\|e_t\|^2$ satisfies
\begin{equation}
\Phi_t \;\le\; (\rho^*)^t\, \Phi_0 \;+\; \frac{C_\sigma\,(\eta_a + \eta_c)\,\sigma^2}{1-\rho^*}.
\label{eq:main-bound}
\end{equation}
A valid concrete choice is $c = 4\eta_a L_{a,\phi}^2/\mu_a$ and $K_0 = \lceil \log(8 L_\phi^2 \eta_a^0 G^2 / \mu_a)/\log(1/\rho)\rceil$, yielding $\rho^* = \max(1-\mu_a\eta_a/2,\ 3/4)$, where $L_{a,\phi} = O(\kappa L_p)$ controls the Actor's sensitivity to Critic perturbations through the margin condition.
\end{theorem}

\begin{proof}
The proof has three steps: the Critic tracking error contracts geometrically, the Actor optimality gap descends modulo Critic error, and a weighted Lyapunov function combines the two.

\paragraph{Critic tracking error.}
The map $\theta \mapsto \phi^*(\theta)$ is $L_\phi$-Lipschitz with $L_\phi \le L_{c,\theta}/\mu_c$, by a standard implicit-function argument under quadratic growth (Karimi et al.~2016, Thm.~5). Since $\phi_{t,0} = \phi_t = \phi^*(\theta_{t-1}) + e_{t-1}$, we obtain
\begin{equation*}
\|\phi_{t,0} - \phi^*(\theta_t)\| \;\le\; \|e_{t-1}\| + L_\phi \|\theta_t - \theta_{t-1}\| \;\le\; \|e_{t-1}\| + L_\phi \eta_a G,
\end{equation*}
where the second inequality uses the bounded-gradient assumption. The standard SGD analysis under quadratic growth yields, for the $K$-step inner loop with $\eta_c \le 1/L_c$,
\begin{equation*}
\mathbb{E}\|\phi_{t,K} - \phi^*(\theta_t)\|^2 \;\le\; \rho^K\, \mathbb{E}\|\phi_{t,0} - \phi^*(\theta_t)\|^2 + \frac{\eta_c\sigma^2}{\mu_c}, \qquad \rho := 1-\eta_c\mu_c.
\end{equation*}
Squaring and combining the two displays via $(a+b)^2 \le 2a^2+2b^2$,
\begin{equation}
\mathbb{E}\|e_t\|^2 \;\le\; 2\rho^K\, \mathbb{E}\|e_{t-1}\|^2 + 2\rho^K L_\phi^2 \eta_a^2 G^2 + \frac{\eta_c\sigma^2}{\mu_c}.
\label{eq:e-rec}
\end{equation}
Choosing $K_0$ so that $2\rho^{K_0} \le 1/2$ (this is the $K_0$ stated in the theorem),
\begin{equation}
\mathbb{E}\|e_t\|^2 \;\le\; \tfrac{1}{2}\mathbb{E}\|e_{t-1}\|^2 + \underbrace{\rho^K L_\phi^2 \eta_a^2 G^2 + \frac{\eta_c\sigma^2}{\mu_c}}_{=:\,\varepsilon_c}.
\label{eq:e-clean}
\end{equation}

\paragraph{Actor descent.}
Let $g_t^* := \nabla_\theta J_a(\theta_t;\phi^*(\theta_t))$ and $g_t := \nabla_\theta J_a(\theta_t;\phi_{t+1})$. The pseudo-label $\hat{a}_k$ flips between $\phi_1$ and $\phi_2$ only on the event $\{|p(s;\phi_1)-\tau|\le L_p\|\phi_1-\phi_2\|\}$, which by the margin condition has probability at most $\kappa L_p\|\phi_1-\phi_2\|$. Combined with the Lipschitzness of BCE in its target argument, this yields
\begin{equation}
\|g_t - g_t^*\| \;\le\; L_{a,\phi}\, \|e_t\|, \qquad L_{a,\phi} = O(\kappa L_p).
\label{eq:grad-perturb}
\end{equation}

By smoothness of $J_a$ in $\theta$ (with the second argument held at $\phi^*(\theta_t)$) and unbiasedness of the stochastic gradient,
\begin{equation*}
\mathbb{E}\, J_a(\theta_{t+1};\phi^*(\theta_t)) \;\le\; J_a(\theta_t;\phi^*(\theta_t)) - \eta_a \langle g_t^*, g_t\rangle + L_a\eta_a^2 \|g_t\|^2 + L_a\eta_a^2 \sigma^2.
\end{equation*}
Substituting $g_t = g_t^* + (g_t-g_t^*)$, using Cauchy–Schwarz and Young's inequality together with \eqref{eq:grad-perturb}, and applying the PL condition $\|g_t^*\|^2 \ge 2\mu_a V_t$:
\begin{equation*}
\mathbb{E}\, J_a(\theta_{t+1};\phi^*(\theta_t)) \;\le\; J_a(\theta_t;\phi^*(\theta_t)) - \mu_a\eta_a V_t + 2\eta_a L_{a,\phi}^2 \mathbb{E}\|e_t\|^2 + L_a\eta_a^2 \sigma^2,
\end{equation*}
provided $\eta_a \le 1/(4L_a)$. Shifting the second argument from $\phi^*(\theta_t)$ to $\phi^*(\theta_{t+1})$ contributes only $O(\eta_a^2)$ by the implicit-function bound and a standard envelope argument; folding this into the noise term gives
\begin{equation}
\mathbb{E} V_{t+1} \;\le\; (1-\mu_a\eta_a)\, V_t + 2\eta_a L_{a,\phi}^2\, \mathbb{E}\|e_t\|^2 + C_1\eta_a^2 \sigma^2.
\label{eq:V-rec}
\end{equation}

\paragraph{Lyapunov function.}
Set $\Phi_t := V_t + c\, \mathbb{E}\|e_t\|^2$ with $c = 4\eta_a L_{a,\phi}^2/\mu_a$. Combining \eqref{eq:e-clean} and \eqref{eq:V-rec},
\begin{equation*}
\mathbb{E}\Phi_{t+1} \;\le\; (1-\mu_a\eta_a) V_t + \big(2\eta_a L_{a,\phi}^2 + \tfrac{c}{2}\big)\, \mathbb{E}\|e_t\|^2 + C_\sigma(\eta_a+\eta_c)\sigma^2.
\end{equation*}
With our choice of $c$, the coefficient on $\mathbb{E}\|e_t\|^2$ equals $\tfrac{c(1+\mu_a)}{2} \le \tfrac{3c}{4}$ (after a harmless rescaling so that $\mu_a \le 1/2$). Hence
\begin{equation*}
\mathbb{E}\Phi_{t+1} \;\le\; \rho^*\, \mathbb{E}\Phi_t + C_\sigma(\eta_a+\eta_c)\sigma^2, \qquad \rho^* := \max\big(1-\tfrac{\mu_a\eta_a}{2},\ \tfrac{3}{4}\big),
\end{equation*}
and iterating yields \eqref{eq:main-bound}.
\end{proof}

\paragraph{Interpretation.}
The bound decomposes joint convergence into two coupled contractions: the Critic tracks $\phi^*(\theta_t)$ at geometric rate $\rho^K$ per outer step, and the Actor descends in $J_a$ at rate $1-\mu_a\eta_a$ modulo a perturbation proportional to the Critic's tracking error. The two-timescale schedule $K=8$ ensures $\rho^K$ is small enough that the Critic does not bottleneck the Actor. The margin condition enters only through the constant $L_{a,\phi}=O(\kappa L_p)$: a sharper Router output (smaller $\kappa$) means small Critic perturbations rarely flip the binarized pseudo-label, so the Actor is robust to a noisy or undertrained Critic. At stationarity,
\begin{equation*}
\limsup_t \mathbb{E} V_t \;\le\; \frac{2L_{a,\phi}^2}{\mu_a} \cdot \limsup_t \mathbb{E}\|e_t\|^2 + O(\eta_a\sigma^2),
\end{equation*}
which formalizes the qualitative claim that \emph{the Actor is no worse than the Critic, up to a constant determined by the margin and PL parameters}. We do not claim global convergence, nor that the equilibrium matches a true oracle policy—the latter would require $J_c$ being a faithful surrogate for translation quality.

\section{Extended Experimental Details}
\label{app:exp_details}

In this section, we provide comprehensive details regarding the training configurations, inference setup, and baseline implementations to facilitate reproducibility. Detailed hyperparameter configurations will be made publicly available alongside our open-sourced code.

\subsection{Load Balancing Loss}

To prevent routing collapse in our two-expert Refiner architecture and maintain numerical stability, we adopt a lightweight, combined load-balancing loss. We first define a squared soft margin penalty, $\mathcal{M}(v; a, b)$, which continuously penalizes a variable $v$ when it deviates from a target interval $[a, b]$:
\begin{equation}
    \mathcal{M}(v; a, b) = 
    \begin{cases} 
    (a - v)^2, & \text{if } v < a \\
    (v - b)^2, & \text{if } v > b \\
    0, & \text{otherwise}
    \end{cases}
\end{equation}
This squared formulation ensures smooth, differentiable gradients at the boundaries. 

Given a batch of $N$ tokens, let $x_i$ denote the raw routing logit for the $i$-th token, and $p_i = \sigma(x_i)$ be its probability of being routed to the first expert. Let $\bar{p} = \frac{1}{N}\sum_{i=1}^{N} p_i$ represent the batch-averaged routing probability. Our overall balancing loss, $L_{bal}$, simultaneously regularizes the micro-level logits to prevent Sigmoid saturation and constrains the macro-level routing probabilities to ensure equitable expert utilization:
\begin{equation}
    L_{balance} = \frac{1}{N} \sum_{i=1}^{N} \mathcal{M}(x_i; -5, 5) + \mathcal{M}(\bar{p}; 0.3, 0.7)
\end{equation}
 By softly bounding individual logits within $[-5, 5]$ and the average routing probability within $[0.3, 0.7]$, the model mathematically guarantees statistical load balancing while preserving token-level routing flexibility.

\subsection{Training and Inference Details}

\paragraph{Hardware and Environment} 
All models were trained and evaluated on a single computation node equipped with eight NVIDIA A100 (80GB) GPUs. Our implementation is built upon the PyTorch framework and PyTorch Lightning. To optimize the memory footprint during training, we integrated DeepSpeed ZeRO Stage 2\cite{rasley2020deepspeed} alongside gradient checkpointing. 

\paragraph{Optimization and Scheduling}
Across all training stages, we employed the AdamW\cite{loshchilovdecoupled} optimizer with $\beta_1 = 0.9$, $\beta_2 = 0.98$, and a weight decay of $0.001$. The learning rate was governed by a linear decay schedule following a 500-step linear warmup. During the parameter-efficient fine-tuning of the Speech LLM backbone, LoRA was applied to all linear modules (excluding word embeddings and the final projection layer) with a rank of $r=16$ and an alpha parameter of $\alpha=32$. The peak learning rate was set to $1.0 \times 10^{-4}$ for Stages 1 and 2, and $2.0 \times 10^{-5}$ for Stage 3. Additionally, the training batch size per GPU was dynamically bounded to a maximum of 30 seconds of audio. Training of each stage takes about 8 hours.

\paragraph{Inference Setup}
For sentence-level inference, we utilized Voice Activity Detection (VAD) exclusively to truncate initial silence at the beginning of audio clips. For document-level inference, VAD was applied solely to filter out excessively long non-speech pauses (e.g., $> 2$ seconds) to prevent severe distribution mismatches with the training data. Acoustic features were continuously extracted using a 400ms window. The KV cache for both the acoustic prefix and the generated text tokens was dynamically maintained and incrementally updated. To minimize inference latency, we adopted a greedy decoding strategy coupled with a windowed 4-gram no-repeat logits processor to prevent degenerative repetitions. All reported latency metrics—for both our proposed method and the baselines—were computed at the word level (or character level for CJK languages).

\subsection{Baseline Implementation Details}
\label{sec:adibaseline}

To ensure a fair and rigorous comparison, we carefully configured all baselines according to their official implementations or API specifications. Both the proposed method and the baselines were evaluated within a unified framework, closely mirroring the official SimulEval protocol. Greedy decoding was uniformly applied across all baseline models.

\begin{itemize}[leftmargin=0.5cm]
    \item \textbf{Seamless Streaming \cite{communication2023Seamless}:} We utilized the official \texttt{seamless-streaming} checkpoint. Its read-write strategy is governed by the native EMMA policy, and we generated the quality-latency Pareto frontier by adjusting the EMMA decision threshold. The audio input stride was set to 400ms, which is identical to our method.
    
    \item \textbf{AlignATT \cite{papi2023AlignAtt} \& StreamATT \cite{papi2024streamatt}:} We followed the implementations of AlignATT and StreamATT provided in the InfiniSST repository\footnote{\url{https://github.com/LeiLiLab/InfiniSST/blob/release/agents/alignatt.py}}, adapting them to operate on top of the original Qwen2.5-Omni Model. We varied the frame boundary to trace the quality-latency trade-off. We have compared extracting attention weights from the middle layer, the final layer, and averaging over layers, showing that an extract from the middle layer yields slightly better results. For StreamATT, the context window was restricted to 40 text tokens, with the retained acoustic history bounded between 10 and 30 seconds. Infinite repetition is hard to avoid in StreamATT, even with the no-repeats logits processor. So we set repetition\_penalty=1.2 in StreamATT.

    \item \textbf{StreamUni \cite{guo2025streamuni}:} We evaluated this baseline using its official checkpoints and code. We use chunk size=640ms and wait-k=3. 
    
    \item \textbf{InfiniSST \cite{ouyang2025infinisst}:} We evaluated this baseline using its official checkpoints\footnote{\url{https://github.com/LeiLiLab/InfiniSST}}. Latency was modulated by varying the input chunk size from 960ms to 4800ms. Since we observed that these checkpoints exhibited degenerate performance during sentence-level evaluations, we only report their document-level results.
    
    \item \textbf{Seed LiveInterpret 2.0:} This model was evaluated via the commercial Volcengine WebSocket API.\footnote{\url{https://www.volcengine.com/docs/6561/1756902}}. To minimize network-induced variance, we ensured the network Round Trip Time (RTT) remained strictly below 20ms during all queries. Due to budget constraints, this API was evaluated exclusively on the FLEURS test set.
\end{itemize}

\begin{figure}[htbp]
     \centering
     \begin{subfigure}[b]{0.8\textwidth}
         \centering
         \includegraphics[width=\textwidth]{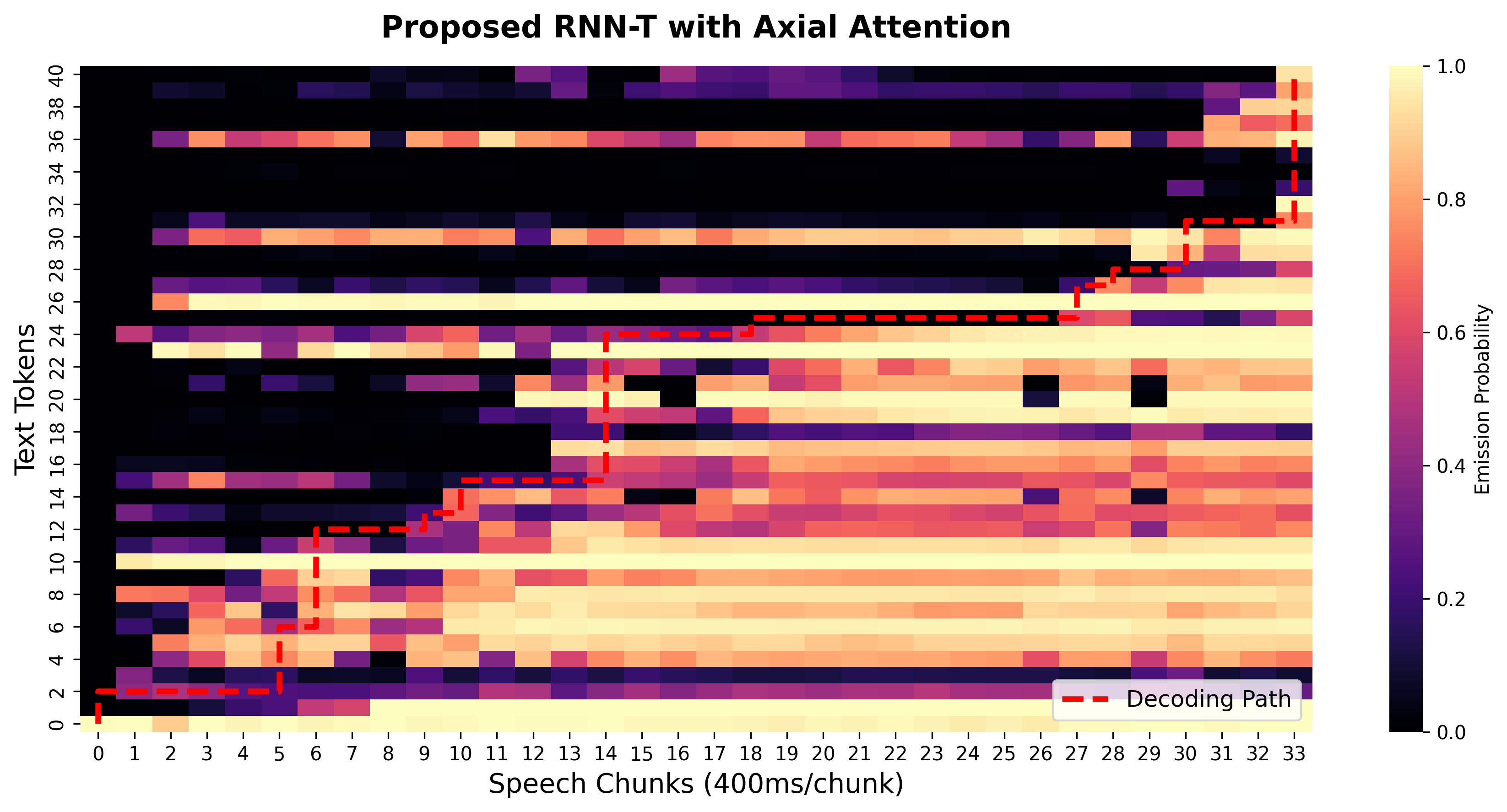}
         \caption{Proposed RNN-T}
         \label{fig:rnnt_p}
     \end{subfigure}
     \vspace{3mm} 
     \begin{subfigure}[b]{0.8\textwidth}
         \centering
         \includegraphics[width=\textwidth]{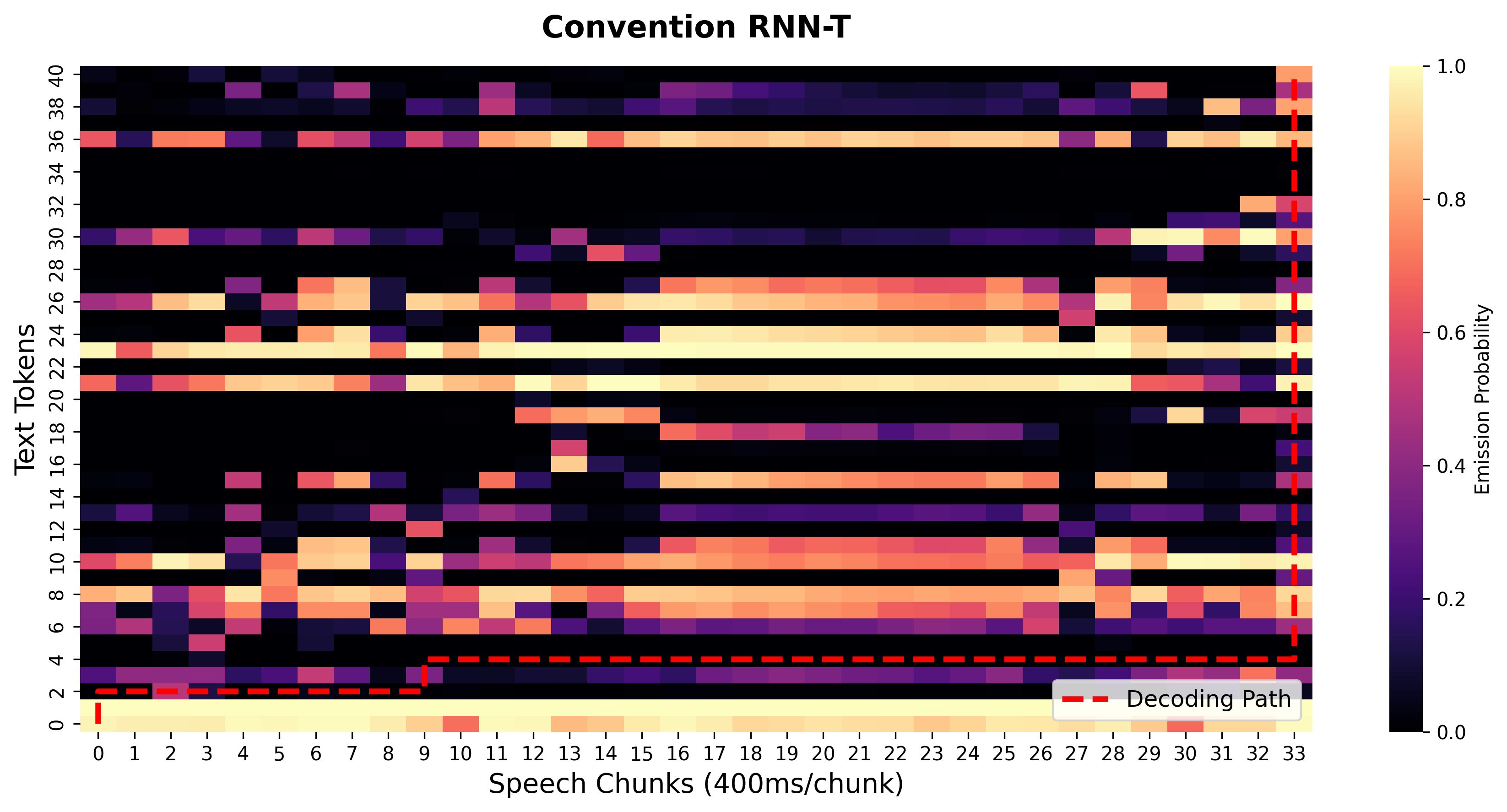}
         \caption{Conventional RNN-T}
         \label{fig:rnnt_c}
     \end{subfigure}
     
     \caption{Comparison of emission probabilities and decoding paths. The proposed RNN-T with axial attention (a) successfully maintains an incremental, monotonic alignment for simultaneous translation. In contrast, the conventional RNN-T (b) fails to emit tokens timely during the speech stream, causing its decoding path to delay and degenerate into an offline translation behavior at the end of the utterance.}
     \label{fig:rnnt_case}

\end{figure}

\section{Additional Experiments}
\label{app:additional_exp}

\subsection{Compatibility with RL-based Policy Optimization}
\label{app:grpo}

CRESiST is an SFT-time policy-discovery framework: translation is still learned by supervised maximum likelihood on the offline corpus, while the read-write policy is discovered by the Actor-Critic loop. It is therefore complementary to, rather than a substitute for, RL-based policy optimization, and the resulting model is a valid starting point for subsequent RL. To verify this, we apply GRPO on top of CRESiST following the reward design of HPO \cite{ouyang2026hierarchical}. For reference, we apply the same recipe to an InfiniSST-style SFT baseline trained on the same Qwen2.5-Omni backbone with RNN-T aligned data (the ``Direct SFT'' setting of Section~\ref{sec:ablations}). CRESiST+GRPO is trained for 700 steps on the multilingual corpus, and SFT+GRPO for 1000 steps on En-Zh. Table~\ref{tab:grpo} reports document-level En-Zh results in the streaming w/o segmentation setting.

\begin{table}[htbp]
\centering
\caption{Document-level En-Zh results (streaming w/o segmentation, XCOMET-XXL $\uparrow$ / LongYAAL $\downarrow$) with and without GRPO post-training. SFT denotes an InfiniSST-style supervised baseline on the same backbone.}
\label{tab:grpo}
\begin{tabular}{lcccc}
\toprule
 & \multicolumn{2}{c}{ACL 60/60} & \multicolumn{2}{c}{MuST-C} \\
\cmidrule(lr){2-3} \cmidrule(lr){4-5}
System & COMET & LongYAAL & COMET & LongYAAL \\
\midrule
SFT (InfiniSST-style) & 68.43 & 2.27 & 66.16 & 2.34 \\
SFT + GRPO (HPO recipe) & 73.64 & 2.15 & 69.71 & 2.15 \\
CRESiST & 79.45 & 1.49 & 73.73 & 1.14 \\
CRESiST + GRPO & \textbf{80.04} & \textbf{1.37} & \textbf{78.47} & 1.23 \\
\bottomrule
\end{tabular}
\end{table}

Three observations follow. (1) CRESiST is fully compatible with RL: GRPO further improves it on both benchmarks (e.g., +4.7 COMET on MuST-C) without hurting latency. (2) RL does not compensate for a weak SFT base: CRESiST without any RL already outperforms SFT+GRPO by a clear margin in both quality and latency, so SFT-time policy learning remains meaningful even when RL post-training is available. (3) Our GRPO gains are smaller than those reported in the HPO paper, which we attribute to the stronger starting point. We also note that after GRPO post-training the model streams raw audio stably without the VAD pause filtering described in Appendix~\ref{app:exp_details}.

\subsection{The RNN-T Actor as an Inference Model}
\label{app:rnnt_infer}

A natural question is whether the self-enhancement loop improves the RNN-T Actor enough for it to serve as the final inference model on its own, which would also quantify how much of the final performance is contributed by the Speech LLM. Our Actor is itself a complete simultaneous translation model trained on the same data (Whisper-medium encoder, Qwen3-0.6B predictor with axial attention, and a joint network), so we decode it directly with streaming greedy decoding (blank = Read, non-blank = Write), both before (Stage 1) and after (Stage 3) reverse distillation. Table~\ref{tab:rnnt_infer} reports sentence-level results on FLEURS.

\begin{table}[htbp]
\centering
\caption{Using the RNN-T Actor directly as the inference model on FLEURS (XCOMET-XXL $\uparrow$ / LAAL $\downarrow$), before and after reverse distillation, compared with the full CRESiST Speech LLM.}
\label{tab:rnnt_infer}
\begin{tabular}{lcccc}
\toprule
 & \multicolumn{2}{c}{En-Zh} & \multicolumn{2}{c}{Zh-En} \\
\cmidrule(lr){2-3} \cmidrule(lr){4-5}
Inference model & COMET & LAAL & COMET & LAAL \\
\midrule
RNN-T Actor, before distillation & 43.7 & 3.20 & 55.5 & 2.29 \\
RNN-T Actor, after distillation & 44.1 & 3.02 & 55.1 & \textbf{1.52} \\
CRESiST (Speech LLM) & \textbf{80.5} & \textbf{1.78} & \textbf{88.6} & 1.54 \\
\bottomrule
\end{tabular}
\end{table}

The results confirm that the RNN-T is effective as an Actor but unsuitable as the final translator. Reverse distillation transfers the \emph{policy}, not the quality: the Actor's COMET is essentially unchanged, while its latency drops markedly (Zh-En LAAL 2.29s $\rightarrow$ 1.52s, close to that of the Speech LLM), showing that its read/write behavior has been aligned with the Critic. Its translation quality, however, remains 33--37 XCOMET points below the Speech LLM; the RNN-T decides reliably \emph{when} to translate, but its 0.6B predictor lacks the linguistic and world knowledge needed for \emph{what} to generate, with named-entity hallucinations being the dominant failure mode. The RNN-T is also an inefficient inference model: each step evaluates the joint network along an entire row or column of the $(T \times U)$ lattice, so its decoding cost grows with the product of the audio and text lengths rather than with the sequence length, and scaling the transducer toward the 7B class would multiply this lattice cost while still lacking pretrained LLM knowledge. These results cleanly separate the two roles in our framework, where the Actor supplies the policy and the Speech LLM supplies the quality, and justify keeping a small RNN-T as the policy explorer paired with a pretrained Speech LLM as the generator.

\section{Case Visualization}

In this section, we present case visualizations to demonstrate the effectiveness of our proposed framework from two distinct perspectives: the internal alignment quality of our speech-text aligner, and the overall end-to-end translation latency.

\subsection{Analysis of RNN-T Alignment Quality}

First, we examine the internal behavior of our proposed RNN-T aligner. Figure \ref{fig:rnnt_case} visualizes the emission probabilities and the resulting decoding paths for both the proposed RNN-T (with axial attention) and a conventional RNN-T on a specific translation case. 

As explicitly shown in Figure \ref{fig:rnnt_c}, the conventional RNN-T struggles to confidently predict the necessary text tokens during the ongoing speech stream. Due to this failure in timely token emission, it cannot establish a proper incremental alignment. Consequently, the decoding path is forced to delay the majority of its predictions until the very end of the speech input (around chunk 33), effectively degenerating the system into an offline translation paradigm.

In contrast, our proposed RNN-T (Figure \ref{fig:rnnt_p}) overcomes this critical limitation. It yields significantly sharper and more concentrated emission probabilities, successfully predicting tokens as the speech stream unfolds. This distinct, high-quality monotonic decoding path demonstrates that our approach effectively maintains the incremental alignment required for true simultaneous translation, avoiding the offline degradation seen in the conventional baseline.

\begin{figure}[htbp]
    \centering
    \includegraphics[width=1\linewidth]{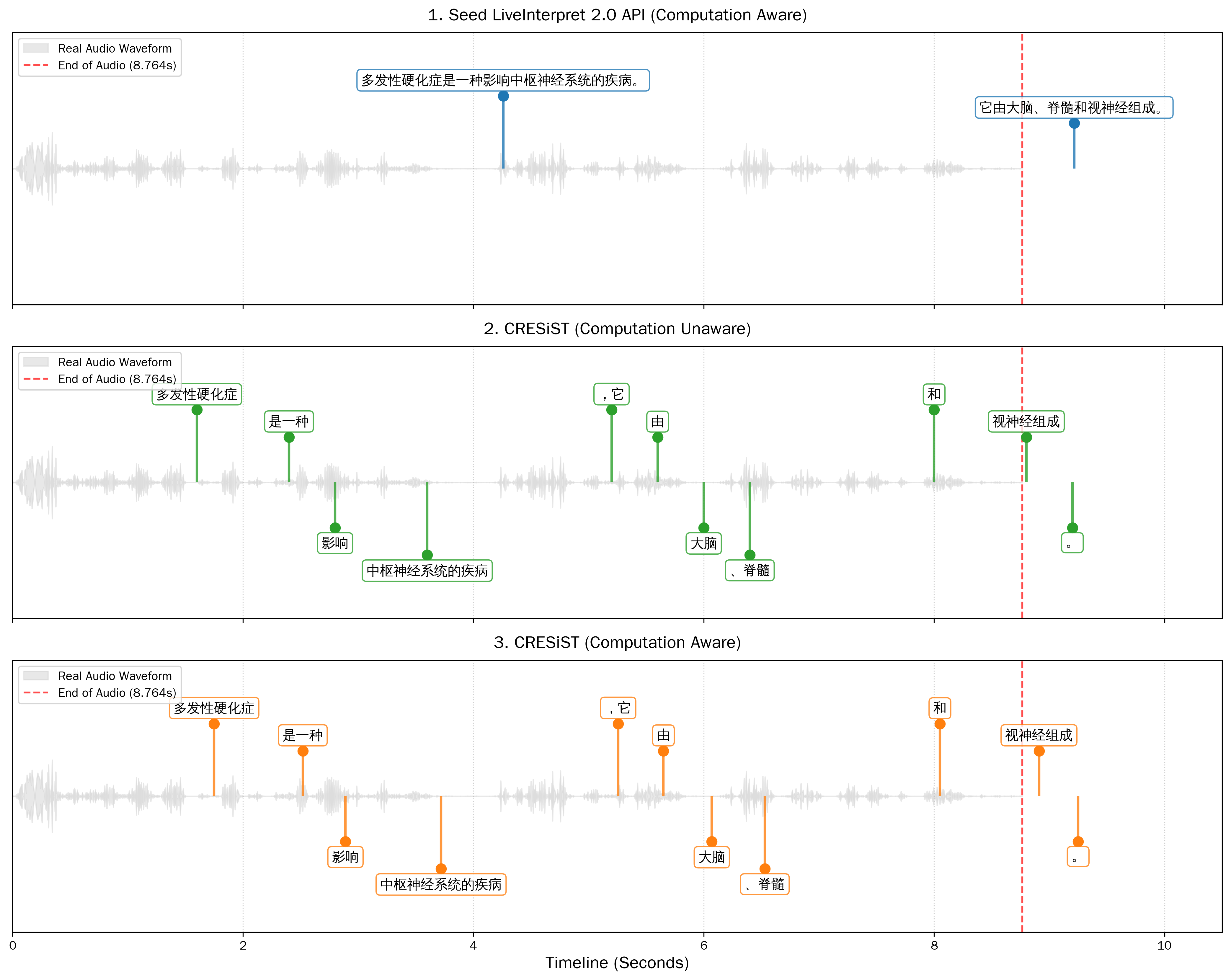}
    \caption{Translation timing visualization for a sample from the FLEURS EN-ZH test set. The timeline illustrates the latency advantage of our model, which performs fine-grained, word-level adaptive translation, in contrast to the coarse-grained, clause-level buffering strategy employed by the Seed LiveInterpret 2.0 API.}
    \label{fig:latency_case}
\end{figure}

\subsection{End-to-End Translation Timing Comparison}

Furthermore, we visualize the end-to-end translation timing of a specific case from the FLEURS EN-ZH test set to compare our approach directly with the Seed LiveInterpret 2.0 API. 

As illustrated in Figure \ref{fig:latency_case}, Seed LiveInterpret 2.0 exhibits a strong preference for coarse-grained, clause-level translation, often waiting for larger chunks of speech context to accumulate before generating an output. Conversely, our proposed model performs fine-grained, adaptive translation at the word level. This allows our model to execute more immediate and flexible read/write decisions, thereby effectively reducing translation latency while maintaining high generation quality.

\section{Assets and Licenses}
\label{app:assets_licenses}

Table \ref{tab:assets} details the existing assets (including datasets, models, and codebases) utilized in this work, alongside their licenses and reference information. 

\begin{table}[htbp]
\centering
\caption{Summary of existing datasets, models, and codebases used in this paper.}
\label{tab:assets}
\resizebox{\textwidth}{!}{
\begin{tabular}{lllll}
\toprule
\textbf{Asset Name} & \textbf{Type} & \textbf{License} & \textbf{Reference}&\textbf{URL}\\
\midrule
\multicolumn{5}{c}{\textit{Datasets}}\\
Emilia & Dataset & CC BY 4.0& \cite{emilia}  &\url{https://huggingface.co/datasets/amphion/Emilia-Dataset}\\
WenetSpeech & Dataset & CC BY 4.0 & \cite{zhang2022wenetspeech}  &\url{https://www.openslr.org/121/}\\
GigaSpeech & Dataset & Apache 2.0 & \cite{chen2021gigaspeech}  &\url{https://github.com/SpeechColab/GigaSpeech}\\
FLEURS & Dataset & CC BY 4.0 & \cite{conneau2023fleurs}  &\url{https://huggingface.co/datasets/google/fleurs}\\
CoVoST 2 & Dataset & CC BY-NC 4.0 & \cite{wang2021covost}  &\url{https://huggingface.co/datasets/facebook/covost2}\\
MuST-C & Dataset & CC BY-NC-ND 4.0 & \cite{di2019must}&\url{https://opendatalab.com/OpenDataLab/MuST-C}\\
ACL 60/60 & Dataset & CC BY 4.0& \cite{salesky-etal-2023-evaluating}&\url{https://aclanthology.org/attachments/2023.iwslt-1.2.dataset.zip}\\
\midrule
\multicolumn{5}{c}{\textit{Models}}\\
Qwen2.5-omni-7B & Model & Apache 2.0 & \cite{xu2025qwen25omnitechnicalreport}  &\url{https://huggingface.co/Qwen/Qwen2.5-Omni-7B}\\
Qwen3-0.6B & Model & Apache 2.0 & \cite{yang2025qwen3technicalreport}&\url{https://huggingface.co/Qwen/Qwen3-0.6B}\\
Qwen3-235B-A22B-Instruct & Model & Apache 2.0 & \cite{yang2025qwen3technicalreport}&\url{https://huggingface.co/Qwen/Qwen3-235B-A22B-Instruct-2507}  \\
Whisper-medium & Model & Apache 2.0 & \cite{radford2023robust}  &\url{https://github.com/openai/whisper}\\
Seamless Streaming & Model& CC BY-NC 4.0 & \cite{communication2023Seamless}  &\url{https://huggingface.co/facebook/seamless-streaming}\\
StreamUni & Model & Apache 2.0 & \cite{guo2025streamuni}&\url{https://huggingface.co/ICTNLP/StreamUni-Phi4}  \\
InfiniSST & Model & MIT & \cite{ouyang2025infinisst}&\url{https://github.com/LeiLiLab/InfiniSST}  \\
\midrule
\multicolumn{5}{c}{\textit{Codebases \& Frameworks \& Metrics}}\\
PyTorch & Framework & BSD-3& &\url{https://pytorch.org/}\\
PyTorch Lightning& Framework & Apache 2.0& &\url{https://github.com/Lightning-AI/pytorch-lightning}\\
DeepSpeed & Codebase & Apache 2.0& \cite{rasley2020deepspeed}  &\url{https://github.com/deepspeedai/deepspeed}\\
OmniSTEval & Codebase & Apache 2.0& \cite{polak2025better}&\url{https://github.com/pe-trik/OmniSTEval/tree/main}  \\
XCOMET-XXL & Metric & CC-BY-NC-SA 4.0 & \cite{guerreiro-etal-2024-xcomet}  &\url{https://huggingface.co/Unbabel/XCOMET-XXL}\\
\bottomrule
\end{tabular}
}
\end{table}

\section{Broader Impacts}
Our proposed CRESiST framework significantly advances simultaneous speech translation, which has the positive potential to break down cross-lingual barriers, facilitate global communication, and improve accessibility for non-native speakers or the hearing-impaired. However, we must also acknowledge potential negative societal impacts. First, like many LLM-based systems, our model may exhibit performance disparities across different demographics, potentially showing bias against non-standard accents or low-resource dialects. Relying on such systems in high-stakes scenarios (e.g., medical or legal settings) without human oversight could lead to critical misunderstandings. Furthermore, the extensive computational resources required for training such omni-modal LLMs contribute to a substantial carbon footprint. We encourage future research to address these fairness and efficiency challenges to ensure safe and equitable deployment.



\end{document}